\documentclass[10pt, conference, compsocconf]{IEEEtran}

\usepackage[utf8]{inputenc} 
\usepackage[T1]{fontenc}    
\usepackage{hyperref}       
\usepackage{url}            
\usepackage{booktabs}       
\usepackage{amsfonts}       
\usepackage{nicefrac}       
\usepackage{microtype}      

\usepackage{textgreek}
\usepackage{amsfonts,amsmath}
\usepackage{amsthm}
\usepackage{algpseudocode}
\usepackage{graphicx}
\usepackage{balance}
\usepackage{algorithm}
\usepackage{tabularx,ragged2e,booktabs,caption}
\usepackage[mathscr]{euscript}
\usepackage{amsbsy}
\usepackage{mathtools}
\usepackage{notes}

\newcommand{\Ber}{\textup{Ber}}

\begin{document}
\title{CODA: Constructivism Learning for Instance-Dependent Dropout Architecture Construction}

\author{\IEEEauthorblockN{Xiaoli Li}
\IEEEauthorblockA{
Ainstein AI\\
Email: xiaoli.li@ainstein.ai}

}

\maketitle

\begin{abstract}
Dropout is attracting intensive research interest in deep learning as an efficient approach to prevent overfitting. 
Recently incorporating ``structural'' information when deciding which units to drop out produced
promising results comparing to methods that ignore the structural information.  However, a major issue of the existing work is that it failed to
differentiate among instances when constructing the dropout architecture. This
can be a significant deficiency for many applications.
To solve this issue, we propose \textbf{Co}nstructivism learning for instance-dependent \textbf{D}ropout \textbf{A}rchitecture (CODA), which is inspired from a philosophical theory, 
constructivism learning. Specially, based on the theory we have designed a better drop out technique, Uniform Process Mixture Models, using a Bayesian nonparametric method Uniform process.
We have evaluated our proposed method on $5$ real-world datasets and compared the
performance with other state-of-the-art dropout techniques. The experimental
results demonstrated the effectiveness of CODA. 
\end{abstract}


\section{Introduction}
Dropout is attracting intensive research interest in deep learning as an efficient approach to prevent overfitting \cite{hinton12}. In the training phase, for each mini-batch, dropout works by randomly omitting some units from the original deep neural network to create a sub-network. In the testing phase, dropout simply computes the average of all the explored subnetworks. Since there is an exponential number of possible sub-networks for a given neural network, it is impractical to explore all of them and then perform model averaging. Drop-out circumvents the problem by adding a regularization that all subnetworks must share the same weights on any shared nodes. With the constraint, the total number of weights need to be trained is still quadratic (assuming a fully connected network) to the number of nodes in the network. The power of dropout for overfitting prevention is attributed primarily to two factors: model averaging with bagging and model regularization. Both reduce model variance.

To design better dropout schemes, a new research thread of incorporating ``structural'' information in deciding units to drop out has produced promising results. For example the work in \cite{tompson15} proposed to drop out simultaneously all the units belonging to a feature map in a convolutional network. Li et al. \cite{li17b} and Neverova et al. \cite{neverova16} utilized modality related information when making drop out decisions. Murdock
et al. \cite{murdock16} developed a method, Blockout, to group the units of a network into clusters which are learned from the data  and dropping out randomly selected nodes becomes dropping out clusters. These methods have obtained better empirical performance in various applications.

A major issue of the aforementioned work is that existing work constructed and applied dropout architectures (subnets) to each instance independently. We call such a strategy \textbf{instance-independent} dropout where ``instance-independent'' refers to the fact the dropout architecture selection for an instance is not affected by the dropout architecture assignments of other instances. Instance-independent dropout failed to differentiate and thus does not consider the relationship among instances. This can be a significant deficiency. For example, in restaurant review prediction where we have studied in our experimental study,  when we predict the ratings given by consumers to different restaurants, consumers may weight the features of a restaurant differently in different activities, such as banquets or dates. Thus a neural network is more likely to achieve better performance if it has the capability to differentiate among instances and construct different dropout architectures for the instances belonging to different activities so that varying weights can be given to the features. 

To address the deficiency of random dropout, we propose \textbf{Co}nstructivism learning for instance-dependent \textbf{D}ropout \textbf{A}rchitecture (CODA), which is inspired from a philosophical theory regarding human learning,  constructivism learning \cite{piaget85, li17a}. This theory has had wide-ranging impact on human learning theories. The essence of this theory is that  human acquire knowledge  from experiences through two fundamental processes: \textit{assimilation}  and \textit{accommodation}. In assimilation, an experience can be incorporated into a learner's  existing knowledge framework without changing that framework.   In accommodation, new knowledge must be constructed in order to accommodate the experience.

\begin{figure*}
  \centering
   \includegraphics[scale=0.4]{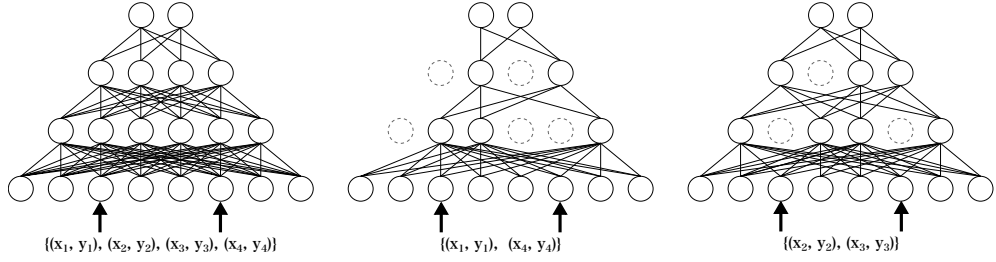}
 \caption{Constructivism Deep Learning. Left: The Network Architecture of A Fully Connected DNN. Middle and Right: Two Different Dropout Architectures. The first dropout architecture is shared by instances $(\vect{x}_1, \vect{y}_1)$ and $(\vect{x}_4, \vect{y}_4)$. The second dropout architecture is shared by instances $(\vect{x}_2, \vect{y}_2)$ and $(\vect{x}_3, \vect{y}_3)$. }\label{fig:cdl}
\end{figure*}

Applying human constructivism learning theory to design better dropout method, for each instance, we believe the key is to decide whether an existing dropout architecture should be used, i.e., assimilation, or a  new dropout architecture should be constructed, i.e., accommodation. We illustrate the concept of constructivism deep learning in Figure \ref{fig:cdl}, where we have a deep neural network (DNN) with two hidden layers, depicted in the left figure. Given $4$ instances $\{(\vect{x}_1, \vect{y}_1), (\vect{x}_2, \vect{y}_2), (\vect{x}_3, \vect{y}_3), (\vect{x}_4, \vect{y}_4)\}$, we further assume that $\{(\vect{x}_1, \vect{y}_1)$ and $(\vect{x}_4, \vect{y}_4)\}$ are similar; that $\{(\vect{x}_2, \vect{y}_2)$ and $(\vect{x}_3, \vect{y}_3)\}$ are similar; and that $\{(\vect{x}_2, \vect{y}_2), (\vect{x}_3, \vect{y}_3) \}$ is quite different from $\{(\vect{x}_1, \vect{y}_1), (\vect{x}_4, \vect{y}_4)\}$. Then for the first instance $(\vect{x}_1, \vect{y}_1)$, a dropout architecture, depicted in the middle figure, is constructed and used. For the second instance, since it is quite different from the first instance, accommodation happens and a new dropout architecture, depicted in the right figure, is constructed for it. For the instance $(\vect{x}_3, \vect{y}_3)$, it triggers the assimilation process, sharing the same dropout architecture with $(\vect{x}_2, \vect{y}_2)$. It is the similar situation for $(\vect{x}_4, \vect{y}_4)$, which shares the same dropout architecture with $(\vect{x}_1, \vect{y}_1)$.

There are many challenges in adapting human constructivism learning to deep learning. First, we need to decide the set of instances should share the same dropout architecture; Secondly, we need to decide the optimal dropout architecture for  those instances. We opted for Bayesian nonparametric techniques for overcoming those challenges by adopting Uniform Process (UP). UP is a Bayesian nonparametric clustering technique with the property that cluster sizes follow uniform distributions.


The contributions of this paper is as follows:
\begin{itemize}
  \item We have adapted human constructivism learning to deep learning to design an instance-dependent dropout technique. 
  \item We have designed an effective algorithm, Uniform Process Mixture Models (UPMM),  with a customized inference method. 
  \item We have launched a comprehensive experimental study with both synthetic and real-world data sets. 
  Comparing the performance with other state-of-the-art dropout techniques, including Blockout, the experimental results demonstrated the effectiveness of our proposed algorithm. 
\end{itemize}
  

\section{Related Work} \label{sec:relatedwork}
In this section, we review two lines of research which are mostly related to our work, dropout training 
and constructivism learning. 
\subsection {Dropout Training for Deep Neural Networks}
Previous work in dropout training for deep neural networks can be categorized into two groups based on whether 
the dropout architectures are determined without or with considering prior knowledge of structures.

For the methods in the first group, the first study was conducted by Hinton et al. in \cite{hinton12}, 
where hidden units were randomly selected using a fixed dropout rate for all the units. In recent years, different variations of
dropout techniques have been developed by approximating the original dropout
technique \cite{gal16, wang13} or learning adaptive dropout rates through
imposing on different distributions, such as multinominal
distributions \cite{li16}, Bernoulli distributions \cite{srinivas16},  distributions based on input activities \cite{ba13},
or employing variational Bayesian inference methods \cite{kingma15, maeda14,molchanov17}. We notice that Bayesian nonparametric (BNP) techniques have also been used in \cite{gal16}. But in our paper, we employ a different BNP method UP since we aim to accommodate the structural information in data while \cite{gal16} only tried to approximate the classical dropout method. 

To incorporate priori structural information in determining dropout architectures, Tompson et al. \cite{tompson15} developed the SpatialDropout
method for convolutional networks to drop out all the units in a feature map simultaneously so that adjacent pixels in the feature map 
are either all inactive or all active. Neverova et al. \cite{neverova16} employed the modality information to drop out the input form a channel 
to achieve robustness in fusion of multiple modality channels for gesture recognition. Different from utilizing these structural information
specific to some applications, Murdock et al. \cite{murdock16} designed Blockout aiming to construct general dropout architectures by grouping units into different
clusters with learned probabilities. \cite{li17b} extended this idea to multi-modal learning by constructing 
dropout architectures so that subsets of units correspond to individual modalities.

All the aforementioned work has failed to address the issue of learning structured dropout 
where different instances may share different dropout architectures or sub-networks. In our experimental study, we have demonstrated the 
advantage of instance-dependent dropout architecture construction. 

\subsection{Constructivism Learning in Machine Learning}
Constructivism learning \cite{piaget85} provides a comprehensive framework of human cognitive development. It has been exploited for interactive machine learning
\cite{sarkar16} and extensively studied in robotic learning \cite{aguilar17}. A complete survey in this field is beyond the scope of this paper and the interested
reader may refer to \cite{stojanov09} for a detailed discussion. 

In \cite{li17a}, to achieve modeling transparency, constructivism learning has been applied to machine learning by 
taking advantage of Bayesian nonparametric techniques, Dirichlet process mixture models.

Note that in this paper, we adapted constructivism learning to deep learning,  which has not been explored in all the aforementioned studies.

\section{Preliminary}\label{sec:preliminary}
In this section, we first introduce the notations used throughout the paper. Then we give a brief overview
of the Bayesian nonparametric technique, Uniform Process, on which our proposed method is based. 
\subsection{Notations}
For clarity, we introduce the following notations.  We use lowercase letters to
represent scalar values, lowercase letters with bold font to represent vectors
(e.g. $\vect{u}$), uppercase bold letters to represent matrices (e.g.
$\mat{A}$), Greek letters $\{\alpha, \lambda, \gamma,\dots\}$ to represent
scalar parameters. Unless stated otherwise, all
vectors in this paper are column vectors.  $\vect{u}^T$ is the transpose of the
vector $\vect{u}$. We use $[1:N]$ to denote the set $\{1,2,\dots,N\}$.

\subsection{Uniform Process}

Uniform process \cite{jensen08, wallach10}  is a variation of Dirichlet Process \cite{ferguson73}. 
An implicit priori property of DP is
``rich-get-richer''.
Thus the sizes of clusters induced by DP are
often non-uniform, with a few very large clusters and some small clusters.
Compared with DP, the advantage of uniform process is that it exhibits uniform distributions over cluster
sizes. 

The partition of a set of observed instances, $\vect{x}_1, \vect{x}_2, \dots$,
can be sequentially constructed using UP as follows.
Given that $N-1$ instances, $\vect{x}_1, \vect{x}_2, \dots,
\vect{x}_{N-1}$, are partitioned into $K$ clusters, let denote the cluster assignment of
$\vect{x_n}$ using an indicator variable $c_n$. For a new instance $\vect{x}_{N}$, 
it will be either assigned to an existing cluster or a new cluster according to the
following probability:
\begin{align} \label{eq:upc}
p(c_{N}=k | ) = 
\begin{cases}
\frac{1}{K+\alpha} \quad k \le K \\
\frac{\alpha}{K+\alpha} \quad k= K+1
\end{cases}
\end{align}
where $\alpha$ is a concentration parameter. It regulates the probability of
assigning an instance to a new cluster. The higher it is, the more likely a new
cluster will be constructed for a new instance.

\section{Algorithm} \label{sec:algorithm}
In this section, we first formalize the problem of \textbf{CO}nstructivism learning for instance-dependent \textbf{D}ropout \textbf{A}rchitecture construction
(CODA) which we aim to solve. Then we describe the details of our proposed method
using UP of mixture models (UPMM) for CODA. Lastly, we outline the inference method designed for the computation of UPMM.

Before proceeding to the details of algorithm, for convenience, we summarize important notations for CODA
in Table \ref{tb:notations}.

\begin{table}
\centering
\caption{Notations for CODA}\label{tb:notations}
\begin{tabular}{|r|l|}
  \hline
  $N$       &  Total number of instances\\
  $M$       & Total number of units in a neural network \\
  $L$       & Total number of labels \\
  $D$       & dimension of features \\
  $K$       & Total number of architectures \\
  $\vect{x}_n$ & feature vector of the instance $n$ \\
  $\vect{y}_n$ & label for $\vect{x}_n$ encoded as a 1-of-L binary\\
               &   vector \\
  $\vect{z}_n$    &  dropout indicator for instance $n$ \\
  $\vect{z}^*_k$    &  dropout indicator for architecture $k$ \\
  $c_n$    &  architecture indicator for instance $n$ \\
  $\mathcal{N}_k$ & indices of instances assigned to architecture $k$ \\
  $\alpha$  &  Concentration parameter for UP \\
  $G_0$      & Base Distribution for UP                  \\			 
  \hline
\end{tabular}
\end{table}

\subsection{COnstructivism learning for instance-dependent Dropout Architecture construction (CODA)}
Suppose we have a set of instances, denoted as a matrix :
\begin{align}
\mat{X} = [\vect{x}_1; \vect{x}_2; \dots; \vect{x}_N] \nonumber
\end{align}
where each row $\vect{x}_n \in
\reals^{D}$ is a row vector and corresponds to an instance, and their corresponding labels, denoted
as a vector $\vect{y} = [y_1; y_2; \dots; y_N]$, $y_n \in [1:L]$. When a
deep neural network is trained using  $\mat{X}$ and $\vect{y}$, 
the previous proposed dropout methods did not consider the possible structure in
data or evaluate the relationship among instances when making decisions about which units
to drop out. Accordingly, the units in the network are randomly selected to omit
only according to the drop out rates, which may be fixed or adaptively learned from
the data. To overcome this limitation, we propose to use \textbf{CO}nstructivism
learning for instance-dependent \textbf{D}ropout \textbf{A}rchitecture construction (CODA). During the training of a deep neural network, the goal of CODA is to determine:
\begin{enumerate}
\item Which instances should share the same dropout architecture for prediction and what the architecture is?
\item When a new dropout architecture should be constructed? 
\end{enumerate}

The above goal characterizes the critical challenge of CODA, that is to 
recognize assimilation, assigning an instance to an existing dropout architecture
 and accommodation, constructing a new dropout architecture for a instance, which 
 corresponds to two fundamental processes of human constructivism learning. 
The solution therefore we seek to implement CODA must have the capability to address this critical challenge.
Specifically, it first needs to have a mechanicism for 
clustering instances so that the dropout architecture inferred from 
those instances are optimal for the prediction performance of the member 
instances in that cluster. Secondly, it should afford a principled way for constructing
a new dropout architecture when a instance cannot be well fitted by existing dropout architectures, which
implies the complexity of the model, mainly assessed by the number of dropout architectures or the amount 
of knowledge learned by the model, needs to be automatically adaptive to the heterogeneity of the data. 

Bayesian nonparametric (BNP) methods has long standing in the literature of statistical and machine learning. 
One major characteristics of BNP is that it is endowed with infinite-dimensional parameter space so that the 
complexity of model parameters is potentially unbounded and the amount of knowledge captured by the model 
increases with increasing number of instances. Counting on this characteristic, we can devise a model based 
on BNP to handle accommodation, constructing new knowledge for an unseen pattern in data. For assimilation, 
we resort to BNP clustering techniques to decide which instances can share the same dropout architecture, i.e.,
explained by the existing knowledge.  Specially, we adopt uniform process (UP), 
a variation of Dirichlet process, and design a UP of mixture models for CODA, 
for which we present the details in the following section.

\subsection{UP Mixture Models for CODA}
Mixture model based on BNP has been widely considered to be one of the most important method for regression and 
classification problems \cite{bastani16, hannah11, shahbaba09, wade14}. It utilizes local regression or classification models, 
such as linear regression or logistic regression, as 
basic building blocks for instances partitioned into different clusters, where instances in the same cluster
share the same model. The distribution of cluster assignments is determined by a mixing measure, which can be a Dirichlet
process or different variations of DP. Generally, the mixture model based on BNP for data $\mat{X}$ and $\vect{y}$ 
assuming the following form:

\begin{align}
y_n|\vect{x}_n, P \sim f(y|\vect{x}_n, P), \;\;
f(y|\vect{x}, P) = \int F(y|\vect{x}, \Phi) dP(\Phi) \nonumber
\end{align}
where $F$ is formulated by the local model used for each cluster and $P$ is the mixing measure. 

Then for each instance $(\vect{x}_n, y_n), \forall n \in[1:N]$, 
the generative process using DP as the mixing measure takes the form:
\begin{align}
y_n |\vect{x}_n,\Phi_n \sim F(y|\vect{x}_n,\Phi_n), \;\;
\Phi_n|G \sim G,\;\;
G \sim DP(\alpha, G_0) 
\end{align}
where $\alpha$ is a concentration parameter of DP, which regulates how likely a new cluster will be constructed.
$G_0$ is a base distribution for model parameters $\Phi_n$. Due to the almost sure discreteness of $G$, some 
$\Phi$'s will have identical values. Then instances and their corresponding model parameters, $\Phi$'s, 
 form clusters; and instances in the same cluster will share the same $\Phi$. 

In Bayesian nonparametric mixture models for classification or regression, for each cluster of instances, 
we need to determine the model parameters $\Phi$, such as regression coefficients in linear regression. For CODA, 
however, our goal is to select dropout architectures. To this end, we parametrize each cluster-specific model 
with a vector consisting of Bernoulli variables $\vect{z} = [z_1; z_2; \dots; z_M]$, where $M$ is the total 
number of neural units in a DNN. $z_i=0$ if unit i is dropped out from the neural network. 
Through $\vect{z}$, we specify a dropout architecture for a cluster. 
For the mixing measure, we use uniform process, 
a variation of Dirichlet process. Then the model we proposed
for CODA can be described as:
\begin{align} \label{eq:upmm}
G_0 = \prod\limits_{m=1}^{M}\Ber(\theta_m),\;\;
G \sim \textup{UP} (\alpha, G_0) \nonumber \\
\vect{z}_n|G \sim G,  \;\;
y_n|\vect{x}_n, \vect{z}_n, \mathcal{W} \sim f(y|\vect{x}_n, \vect{z}_n, \mathcal{W}) \quad for \; n=[1:N] \nonumber
\end{align}
where we use $\Ber$ to denote Bernoulli distribution and $\theta$'s are parameters of Bernoulli distribution. 
$M$ is the total number of neural units in DNN.
$w_{ij}$ is the weight from unit $i$ to unit $j$ and these two units are not in the same layer of DNN. And we use
$\mathcal{W}$ to denote the set of all $w_{ij}$'s.
$N$ is the total number of instances. Note that for simplicity, 
we assume independence for Bernoulli variables $z$'s. 

The choice of probability form for $y_n$ depends on the type of a neural network and its output. For example, 
for multi-layered neural networks with Gaussian outputs, we may use a multivariate Gaussian for the distribution
of $y_n$. In this paper, we focus on relatively simple neural networks with softmax function as output layers.
We therefore compute the probability of $y_n$ using:
\begin{align}
p(\vect{y}_n | \hat{\vect{y}}_n) = \textup{exp} \left [\sum\limits_{l=1}^{L}y_{n,l}\log{\hat{y}_{n,l}}\right ]
\end{align} 
where $\vect{y}_n$ is generated by encoding $y_n$ as a 1-of-L binary vector.  $\vect{\hat{y}}=[\hat{y}_{n,1};
\hat{y}_{n,2}, \dots, \hat{y}_{n,L}]$ is the output value after propagation of $\vect{x}_n$ through the network.

Similar to the Dirichlet process, $G$ drawn from UP is discrete a.s.. Hence $\vect{z}$'s present ties 
with positive probability. Accordingly, instances are partitioned into different clusters, with the same
$\vect{z}$ being shared by all the instances in the same cluster. Since the dropout architecture is completely
determined by $\vect{z}$, the instances in a cluster will also share the same dropout architecture. This provides 
the model a mechanisim for determining which instances should share a dropout architecture, i.e., assimilation. On the other
hand, from (\ref{eq:upc}) we can observe that given the partitions of $N-1$ instances into $K$ clusters, 
a new instance has a positive probability proportional to $\alpha/(K+\alpha)$ to be assigned to a new architecture,
which enables accommodation. 

For the computation of  (\ref{eq:upmm}),  we need to infer the parameters:
$\Omega = \{\mat{Z}^*, \mathcal{W}\}$. Here we use $\mat{Z}^* = [\vect{z}^*_1, \vect{z}^*_2, \dots, \vect{z}^*_K]$ 
to denote the distinct values of $\vect{z}_n$, $\forall n \in [1:N]$.

\subsection{Computation}
We adapted the method proposed in \cite{lin13} for the computation of UPMM since it is sequential and can be used for
non-conjugate situations, which is the case in our proposed UPMM model. In addition, it allows model parameters
$\mathcal{W}$ to be efficiently updated in mini-batches using stochastic optimization methods.

One major issue of computation of UPMM is the inference of $\mat{Z}^*$ since it is discrete and optimization methods
based on stochastic gradients are infeasible. To solve this issue, we propose a method for updating $\vect{z}^*_{k}$ using
all the instances that share $\vect{z}^*_{k}$ at once instead of updating stochastically by mini-batches.
Although this method may incur more computation time, we found that the efficiency performance is acceptable for the
data sets we used in our experiments.

In the following, we first describe how to assign instances to different architectures. Then we give the details of updating
model parameters $\mathcal{W}$ and $\mat{Z}^*$. Lastly, we present how the model is used for the prediction of
test instances.
\subsection{Update architecture Assignment}
To determine which instances should share the same dropout architecture, that is, which instances should be
partitioned into the same cluster, we introduce latent variables $c_n$ for instance $(\vect{x}_n, \vect{y}_n)$ to indicate the assignment of the architecture. We have $c_n = k$ iff $(\vect{x}_n, \vect{y}_n)$ is assigned to architecture
 $k$. Then the probability of architecture assignment for $(\vect{x}_n, \vect{y}_n)$ given the architecture assignments
 of other instances is as follows:
 \begin{align} \label{eq:updatec}
 \rho_k(c_n=k|\cdot) \propto
 \begin{cases}
 \int_{\vect{z}^*} f(\vect{y}_n | \vect{z}^*, \mathcal{W}) \nu_k(d_{\vect{z}^*}) \quad k \le K\\
 \alpha \int_{\vect{z}^*} f(\vect{y}_n | \vect{z}^*, \mathcal{W}) G_0(d_{\vect{z}^*}) \quad k = K+1
 \end{cases}
\end{align}

Note that our method is different from \cite{lin13} in that we use hard-clustering for each instance.
We choose this strategy due to the following considerations. First, we estimate the
model parameters through a number of iterations while \cite{lin13} only performs one single pass over the data.
Secondly and most importantly, by using hard-clustering, we only need use those instances belonging to architecture $k$
to update architecture-specific parameters $\vect{z}$. With soft-clustering, all the instances need to be used for the
 updating of parameters of each architecture. This may be computationally daunting when inferring from relatively large data sets.

 \textbf{Regularization through Similarity among Instances}. In (\ref{eq:updatec}),
 the assignment of dropout architectures is mainly determined by the prediction
 performance of each architecture. This strategy may raise two issues.
 Firstly, the probability that several architectures have similar prediction
 performance is high. Although each dropout architecture corresponds to a different decision boundary,
 the number of potential decision boundaries that have similar prediction performance for one instance is large.
 Thus it poses challenge in determining which architecture should be used. Secondly, it is likely
 to construct a relatively large number of architectures with a small number of instances assigned to each architecture
 if the prediction performance is used as the only assignment criteria. This may lead to overfitting since it is difficult
 to have a architecture well trained with limited number of instances and the generalization performance will be low.

 To alleviate these two problems, we propose to regularize the architecture assignment based on similarity among instances. Our
 assumption is that similar instances tend to use the same dropout architecture.
 Specially, when making the decision whether an instance $(\vect{x}_n, \vect{y}_n)$ should be assigned to the architecture $k$,
 we also consider the similarity between $\vect{x}_n$ and other instances which have been assigned to architecture
 $k$ in addition to the prediction performance of using architecture $k$. Thus we add an regularization term to (\ref{eq:updatec}) to
 get the following equation:
 \begin{align} \label{eq:updatec-reg}
 \rho_k(c_n=k|\cdot) \propto
 \begin{cases}
 s_k^{\beta_1}(\int_{\vect{z}^*} f(\vect{y}_n | \vect{z}^*, \mathcal{W}) \nu_k(d_{\vect{z}^*}))^{\beta_2}  \quad k \le K\\
 \alpha \int_{\vect{z}^*} f(\vect{y}_n | \vect{z}^*, \mathcal{W}) G_0(d_{\vect{z}^*}) \quad k = K+1
 \end{cases}
\end{align}
where $s_k \in \reals$ is used to denote the similarity between $(\vect{x}_n, \vect{y}_n)$ and other instances assigned to architecture $k$.
$\beta_1, \beta_2 \in \reals$ are regularization parameters. Let denote the set of instances assigned to architecture $k$ as $\mathcal{N}_k$.
To compute $s_k$, we first compute the mean of $\mathcal{N}_k$ using:
\begin{align}
\vect{m}_k = \frac{1}{|\mathcal{N}_k|} \sum\limits_{\vect{x}_i \in \mathcal{N}_k} \vect{x}_i \nonumber
\end{align}
Then $s_k$ is computed based on the distance between $\vect{m}_k$ and $\vect{x}_n$:
\begin{align}
s_k = \textup{exp}(- \|\vect{x}_n-\vect{m}_k\|^2_2) \nonumber
\end{align}

\subsection{Update $\mat{Z}$}
Since $G_0$ and $f(\vect{y}_n | \vect{z}_n, \mathcal{W}) $ are not a conjugate
pair, there exist no closed-form formulas for calculating the posterior
probability of $\vect{z}^*$.
Given the architecture assignments of instances, we can only know that the
posterior probability of $\vect{z}^*$ proportional to the form:
\begin{align} \label{eq:updatez}
\nu_k(d_{\vect{z}^*}) \propto
\begin{cases}
G_0(d_{\vect{z}^*})\prod\limits_{i\in{\mathcal{N}_k}}f(\vect{y}_n | \vect{z}^*, \mathcal{W}) \quad k \le K \\
G_0(d_{\vect{z}^*})f(\vect{y}_n | \vect{z}^*, \mathcal{W}) \quad k = K + 1
\end{cases}
\end{align}
Thus we propose to address this problem using MAP point estimation since $\vect{z}^*$ is discrete and each element
$z^*_m, \forall m \in [1:M]$ in $\vect{z}^*$ will take on either value $1$ or value $0$. Specially, for the estimation
 of $z^*_m$, we fix the values of $z^*_i, \forall i \in [1:M]$ and $i \ne m$, then select the value of $z^*_m$ so that
 (\ref{eq:updatez}) is maximized.

 \textbf{Preventing Local Optimum}. The disadvantage of using MAP point estimation for updating $\mat{Z}$ is that it may
 trap into local optimum. To avoid this, we employ an updating strategy based on the Simulated Annealing (SA) algorithm proposed
 in \cite{locatelli01}. In each iteration, when determine whether the new value of $z^*_m$ should be accepted, there are two
 cases. In the first case, $z^*_m$ will take on the new value if the value of (\ref{eq:updatez}) is larger. In the second case,
 $z^*_m$ will take on the new value with probability $p$ even if the value of (\ref{eq:updatez}) is smaller. Here $p$ is
 calculated as follows:
 \begin{align}
 p = \textup{exp}(\textup{log}\nu^n_k-\textup{log}\nu^o_k)/T) \nonumber
 \end{align}
 where $\nu^n_k$ is calculated
 from (\ref{eq:updatez}) using the new value of $z^*_m$; and $\nu^n_k$ is calculated using the old value of $z^*_m$.
 $T$ is updated in each iteration with $T = \gamma_1(\textup{log}\nu_k)^{\gamma_2}$. Here $\nu_k = \nu^n_k$ if the new value
 is assigned to $z^*_m$, otherwise $\nu_k = \nu^o_k$.
 The intuition behind this strategy is that when $\nu_k$ is far away from the optimal value, the probability of  $z^*_m$ taking on the new value
 is high even if that new value leads to smaller $\nu_k$ so that the parameter space explored by the algorithm will be larger.

\subsubsection{Update $\mathcal{W}$}
To reduce the variance of gradient estimation, we use mini-batches for the updating of $\mathcal{W}$ through backpropagation.
The specific procedure is as follows. In each iteration of training, the training data arrive sequentially in mini-batches.
Given the $b$th mini-batch containing $I$ instances $(\vect{x}_{b,1}, \vect{y}_{b,1})$,
 $(\vect{x}_{b,1}, \vect{y}_{b,1})$,$\dots$, $(\vect{x}_{b,I}, \vect{y}_{b,I})$ we first determine the architecture
 assignments of each instance according to (\ref{eq:updatec-reg}) to get $c_{b,1}, c_{b,2},\dots, c_{b,I}$. Let denote the distinct
 values of  $c_{b,1}, c_{b,2},\dots, c_{b,I}$ as $d_1; d_2;, \dots; d_J$ and the set of instances assigned to
 architecture $d_j$ as $\mathcal{S}_j$,
 then for each architecture $d_j$, we update the weights of that architecture following the same process
 in original dropout training by using $\mathcal{S}_j$, back propogating only through those nodes which are
 kept in the architecture after dropout.

\subsection{Prediction}
The strategy we use for the prediction of a test sample $\vect{x}$ is as follows.
First, we propagate forward through each dropout architecture
 to generate the $K$ output vectors,  $\hat{\vect{y}}_1$, $\hat{\vect{y}}_2$, $\dots$,
$\hat{\vect{y}}_K$. Next we select the maximum element in in each vector to get
$\hat{y}^*_1$, $\hat{y}^*_2$, $\dots$,
$\hat{y}^*_K$ and their corresponding indices, $i_1$, $i_2$, $\dots$,$i_K$, in each output vector.
After have computed the similarity between $\vect{x}$ and $\mathcal{N}_k$, $\forall k \in [1:K]$
to get $s_1$, $s_2$, $\dots$, $s_k$, we assign $\vect{x}$ to the architecture $k$ based on both $\hat{y}^*_1$
and $s_1$. That is, we have the cluster assignment of $\vect{x}$:
\begin{align}
 c=\max\limits_{k}s_k^{\beta_1}(\hat{y}^*_k)^{\beta_2}
\end{align}
and assign the label of $\vect{x}$ to $i_c$.

We summarize the computation procedure in Algorithm \ref{alg:upmm}.
\begin{algorithm}
\caption{CODA using UPMM}
\label{alg:upmm}
\begin{algorithmic}[1]
\State \textbf{Input: }$\mat{X},\mat{Y},numEpochs, numBatches$
\State \textbf{Initialize: }$T\gets 0, V \gets 0$
\For {$t < numEpoths$}
\For {$b < numBatchs$}
\State get $b$th batch of instances $(\mat{X}_b, \mat{Y}_b)$
\For {each instance  $(\vect{x}_{b,i}, \vect{y}_{b,i})$ in $(\mat{X}_b, \mat{Y}_b)$}
\State Assign dropout architecture according to (\ref{eq:updatec})
\EndFor
\State Update $\mathcal{W}$ according to architecture assignments of
\State $(\mat{X}_b, \mat{Y}_b)$
\State $b \gets b+1$
\EndFor
\For {$k$ = $1$ to $K$}
\State Update dropout indicator $\vect{z}^*_k$
\EndFor
\State $t \gets t+1$
\EndFor
\State \textbf{Output: }$\mat{Z}^*,\mathcal{W}$
\end{algorithmic}
\end{algorithm}

\section{Experiments} \label{sec:experiment}
To investigate the performance of our proposed method, we evaluated it on $5$ real-world data sets and compared the results with $2$ other state-of-the-art dropout techniques. In the following, we begin by describing the details of those data sets and methods being compared. Then we present the specific protocol used for the experiments. Lastly, we analyze the experimental results and give a detailed discussion.

\subsection{Data Sets}
The $5$ real-world data sets and $4$ synthetic data sets used 
for our experiments are described in the following.

\subsubsection{Synthetic Data Sets}
The group of synthetic data sets, denoted as SDS1, $\dots$, SDS4, were generated using multi-layer neural networks.
with $U$ units in each hidden layer, where $U \in [25, 50, 75, 100]$. 
We first generate the weights of neural networks from a Normal distribution:
\begin{align}
\vect{w}_{ij} \sim \textup{N}(0, 1)  \nonumber
\end{align}
where $i$ is the index of a unit in layer $h$ and $j$ is the index of a unit in layer $h+1$. Here $h \in [1:H-1]$ and $H$ is the total
number of layers.
We generated the features using a Multivariate Normal Distribution:
\begin{align}
\vect{x}_n \sim \textup{MN}(\vect{m}_k, \mat{\Sigma}_x) \quad for \; n=[1:N] \nonumber
\end{align}
where $\vect{m}_k \in \reals^D$, $k \in [1:3]$. We use $k$ to denote indices of dropout architectures.  And $\mat{\Sigma}_x$ is a diagonal matrix having
$50$'s as its diagonal elements. $D$ is the dimension of a data set and $N$ is the total number of instances in that data set.
For each data set, we constructed $3$ different dropout architectures by randomly and uniformly dropout $50\%$ of the units in each hidden layer.
For each dropout architecture, $2000$ instances were generated from Multivariate Normal distributions using mean $\vect{m}_k$, where $m_{1,d} = 0$,
$m_{2,d} = 5$, and $m_{3,d} = -5$ for $d \in [1:D]$. Here we use $m_{k,d}$ to denote the $d$th element in vector $\vect{m}_k$. After having generated the weights and features,
we propagate forward through the dropout architecture to get the labels.
The details of each data set are summarized in Table \ref{tb:dataset-syn}.

\begin{table}
\centering

\begin{tabular}{|c|c|c|c|c|c|}
 \hline
  Data set      & N & D  & L& U & K \\
 \hline
 SDS1       & 6000    & 50   & 2   & 25  & 3\\
 SDS2       & 6000    & 100  & 2   & 50  & 3 \\
 SDS3       & 6000    & 150  & 2   & 75  & 3 \\
 SDS4       & 6000    & 200  & 2   & 100  & 3 \\
 \hline
\end{tabular}
\caption{Statistics of Synthetic Data Sets. N: Number of Instances, D: Number of features, L: Number of Labels, U: number of hidden units in each hidden layer, K: number of dropout architectures}\label{tb:dataset-syn}
\end{table}

\subsubsection{Real-world Data Sets}
In this section, we introduce the $5$ real-world data sets, 
which are Japan Restaurant data set, Spam E-mail data set, Income data set, Crime data set, and Creditcard data set, which we used for the performance evaluation of different algorithms.

\textbf{Japan Restaurant Data Set}. This data set contains $800$ ratings on 
$69$ restaurants in Japan from $8$ users \cite{oku06}. There are $30$ features, including 
both restaurant attributes and  event related parameters. All the features are used in the experiment.
 The prediction task for this data set is to estimate a user's rating for a
restaurant given the restaurant's attributes and context conditions. 

\textbf{Spam E-mail Data Set}. This data set \cite{lichman13} is composed of $4601$ instances with $57$ features for each instance.
The first $54$ features denote whether a particular word or
character is frequently occurring in an e-mail.  The rest of the features
indicate the length of sequences of consecutive 
capital letters. The prediction task for this data set is to determine whether an e-mail is a spam or not. 

\textbf{Income Data Set}.  The $45222$ instances in the income data set \cite{lichman13} were generated in 1994 from census data of 
the United States. The original data set has $14$ features consisting of both continuous and nominal attributes.
We encoded those categorical features with $C$ unique values
as 1-of-C binary vectors to get $65$ features. The task is to predict whether a citizen's income exceeds fifty thousand dollars per year or not.

\textbf{Crime Data Set}. The original data set consists of $1994$ instances with $128$ features \cite{lichman13}. .
The predicted label is the normalized total number of violent crimes per 100K population. 
In our experiments, we removed those features with missing data and only used the rest $100$ features. 
For the label, we converted it to 1 when it is larger than 0.5, and 0 otherwise.

\textbf{Creditcard Data Set}. This data set provides $30000$ records of credit card clients in Taiwan \cite{yeh2009}. There are $23$ features, containing data about 
clients' payment history and personal information, such as age, gender, and education. The task is to predict whether a client will default payment or not.

%
%

\subsection{Compared Methods}
 For the compared methods, we used fully connected multi-layer deep neural networks (DNN) without dropout as the baseline. 
 In addition, we compared our proposed method with the original dropout method proposed by Hinton et al.
\cite{hinton12} and other $2$ variations of dropout techniques, a very recently developed  sparse variational dropout (sparseVD) method \cite{molchanov17} 
which does not consider the structural information, and the Blockout method \cite{murdock16} which assumes that there exist a predefined number of 
dropout architectures and groups the units accordingly. 
 
\subsection{Experimental Protocol}
\textbf{Network Architecture}. 
In this paper, we focus on multi-layer neural networks. For the neural networks, we used $3$ hidden layers and $20$ units in each layer for all the real-world data sets expect crime data set.
The crime data set has relatively large number of features. Thus $50$ units were used in each hidden layer. 
For synthetic data sets,  we used the same network architectures from which the data were generated. 
For the activation function and output function, we use sigmoid and softmax respectively. 
Accordingly, cross-entropy loss is employed for gradient descent optimization. 
The loss
is defined as:
\begin{align}
-\frac{1}{N}\sum\limits_{n=1}^{N}\sum\limits_{l=1}^{L}y_{n,l}\log{\hat{y}_{n,l}}
\end{align}
where $N$ is total number of instances. $L$ is the total number of labels. $\vect{y}_n$ is generated by encoding the label of the instance $\vect{x}_n$ as a 1-of-L binary vector.  $\vect{\hat{y}}=[\hat{y}_{n,1};
\hat{y}_{n,2}; \dots; \hat{y}_{n,L}]$ is the output value after propagation of $\vect{x}_n$ through the network. Note that all the networks were trained with random initialization. 

\textbf{Model Selection}.
For each data set, we used 50\% the data as training data and the rest as test
data. We tuned model hyper-parameters for each algorithm using 10-fold cross
validation on training data. Once with the best model hyper-parameters, we train a single model on the training data set and evaluate the model on the testing data set. 
We repeat the experiments 10 times to evaluate statistical significance of results. 

\textbf{Model Evaluation Metric}.
We chose F1 score as the performance metric because Creditcard and Crime data sets are rather imbalanced.  

\textbf{Significance Test}. When comparing different methods, we made sure
that those methods were trained using the same training data sets and were
evaluated on the same testing data sets. To evaluate the statistical significance of the difference between different results, we conducted paired student's t test.

\subsection{Experimental Results and Discussion}
We first describe the results of performance evaluation of different methods.
To investigate why CODA can achieve better performance than the classical dropout technique, we also performed a case study using a real-world data set.
We first studied the effectiveness of our proposed optimization method.
 Specially, we investigated the effects of two techniques,
similarity based regularization (SReg) and utilizing utilizing Simulated Annealing for preventing local optimum (SA), on improving the performance of optimization. Then we compared 
the performance of different algorithms using F1 score. 

\subsubsection{Optimization Evaluation}
To see how SReg and SA can affect the effectiveness of optimization, we designed $4$ experiments for each data set as follows. 
As the baseline method, we performed the optimization using neither SReg or SA. Then we use SReg or SA separately for optimization. For
the last experiment, we evaluated the combining effects of SReg and SA on the optimization.  

We show the comparison among different optimization strategies on synthetic data sets in Table \ref{tb:opperf-syn}. We observe consistent
improvement brought by employing SA, SReg, or both on synthetic data sets.  For all the $4$ data sets, we achieved better 
performance when applying SA during the optimization process.  Compared with SA, the advantage of utilizing SReg is more significant.
It outperforms the baseline method with a large margin. By combining SA with SReg, the performance can be further boosted and the difference 
is statistically significant on $3$ data sets. 

The optimization evaluation on real-world data sets is presented in Table \ref{tb:opperf-real}. Although it has slightly worse performance than the baseline method
on Income data set, the utility of applying SA can still be validated on the other $4$ data sets. Especially on Creditcard data set, the performance differs by
more than one order of magnitude. Taking advantage of SReg, we achieved better performance than using SA on $3$ data sets, JapanRestaurant, Income, Creditcard. 
On the other two data sets, it shows an advantage over the baseline method although it performed worse than SA. For the combining of SA and SReg, the performance
is slightly worse than using SA on Crime data set and comparable to SReg on Creditcard data set. But the apparent improvement attained on the first
$3$ data sets, JapanRestaurant, Spam, and Income underlines the importance of using both SA and SReg. Interestingly, despite the undesirable performance of SA
on Income data set, the synergistic effect of combining both SA and SReg on improving optimization is evident.

\begin{table}
\centering
\begin{tabular}{|c|c|c|c|c|c|}
 \hline
  Data set       & Base      & SA       &  SReg & SA+SReg \\
 \hline
 SDS1            & 0.671     & 0.686    &  0.733     & \textbf{0.737}         \\
 SDS2            & 0.614     & 0.642    &  0.712     & \textbf{0.722*}      \\
 SDS3            & 0.590     & 0.613    &  0.703     & \textbf{0.711**}      \\
 SDS4            & 0.622     & 0.635    &  0.686     & \textbf{0.694*}    \\
   
 \hline
\end{tabular}
\caption{Optimization Evaluation  on Synthetic Data Sets. **: statistically significant with $1\%$ significance level; *: statistically significant with $5\%$ significance level.}\label{tb:opperf-syn}
\end{table}

\begin{table}
\centering
\begin{tabular}{|c|c|c|c|c|c|}
 \hline
  Data set         & Base       & SA  &  SReg & SA+SReg \\
 \hline
 Japan Restaurant  & 0.524     & 0.533    &  0.559     & \textbf{0.602**}         \\
 Spam E-mail       & 0.546     & 0.570    &  0.553      & \textbf{0.628*}      \\
 Income            & 0.502     & 0.496    &  0.561         & \textbf{0.603*}    \\
 Crime             & 0.591     & \textbf{0.613}    &  0.600     & 0.590    \\
 Creditcard        & 0.059     & 0.190   &  0.285      & \textbf{0.291}   \\
         
 \hline
\end{tabular}
\caption{Optimization Evaluation on Real-world Data Sets}\label{tb:opperf-real}
\end{table}

\subsubsection{Performance Evaluation}
We compared our proposed method, CODA, with the baseline method, fully connected multi-layer deep neural networks (DNN), and different variations of dropout methods on both
synthetic data sets and real-world data sets, as shown in Table \ref{tb:perf-syn} and Table \ref{tb:perf-real} respectively.

For the synthetic data sets,  dropout surpasses the baseline method narrowly on SDS2 while performance slightly worse on the other $3$ data sets. For sparseVD,
It shows advantageous or comparable performance over DNN and Dropout on all the $4$ data sets.  Compared with other methods, Blockout performs worse
on all the synthetic data sets, with an noticeable sharp decrease on SDS1.  The possible explanation for this result is that there is no constraint enforcing the probabilities of dropout architecture assignments between $0$ and $1$ during the optimization process, which may lead to undesirable effects. For our proposed method, CODA, the advantage over other methods is statistically significant on all the synthetic data sets. 

From the comparison results of different algorithms on the real-world data sets, we observe that Dropout only achieves better performance than DNN on Crime data set. For sparseVD, the performance on Crime data set is comparable to DNN and Dropout despite that it performs much worse on the other $4$ data sets. Compared with sparseVD, Blockout has achieved better or comparable performance on $4$ data sets.  However, it performs significantly worse than other methods on Crime data set. CODA beats other methods with statistical significance level $1\%$ on Japan, Spam, and Income data sets and $5\%$ on Creditcard data set. This result confirms
the advantage of CODA. 

\begin{table*}
\centering
\begin{tabular}{|c|c|c|c|c|c|c|}
 \hline
  Data set         & DNN       & Dropout  &  sparseVD & Blockout & CODA \\
 \hline
 SDS1             & 0.682     & 0.680    &  0.683     & 0.001    & \textbf{0.737**}      \\
 SDS2             & 0.648     & 0.641    &  0.659     & 0.609    & \textbf{0.722**}   \\
 SDS3             & 0.628     & 0.635    &  0.646     & 0.462    & \textbf{0.711**}      \\
 SDS4             & 0.649     & 0.645    &  0.652     & 0.593     & \textbf{0.694**}      \\
 \hline
\end{tabular}
\caption{Model Performance using F1 score with Different Methods on Synthetic Data Sets}\label{tb:perf-syn}
\end{table*}

\begin{table*}
\centering
\begin{tabular}{|c|c|c|c|c|c|c|}
 \hline
  Data set         & DNN       & Dropout  &  sparseVD & Blockout & CODA \\
 \hline
 Japan Restaurant  & 0.531     & 0.396    &  0.133          & 0.154    & \textbf{0.602**}      \\
 Spam E-mail       & 0.533     & 0.363    &  0.284          & 0.400    & \textbf{0.628**}   \\
 Income            & 0.193     & 0.116    &  0.077          & 0.155    & \textbf{0.603**}      \\
 Crime             & 0.582     & 0.604    &  0.594          & 0.014    & \textbf{0.613}      \\
 Creditcard        & 0.182     &0.109     &  0.182          & 0.182     & \textbf {0.291*}  \\
 \hline
\end{tabular}
\caption{Model Performance using F1 score with Different Methods on Real-world Data Sets}\label{tb:perf-real}
\end{table*}

\subsubsection{Case Study}
We conducted a case study on Japan Restaurant data set to investigate how instance-dependent dropout architecture construction can affect the performance of algorithms.
To this end, we analyzed the $2$ dropout architectures, denoted as $d_1$ and $d_2$, constructed by CODA for the data set and noticed a discrepancy between 
the instances assigned to $d_1$ and the ones assigned to $d_2$. It was found that the number of 
instances having the feature, recommended for banquets, denoted as $b$, in $d_1$ is almost twice the number of instances having this feature in $d_2$. Based on this observation,
we hypothesize that the performance can be improved if we split the instances into $2$ groups according to whether they have the feature $b$ or not 
and train $2$ different networks for them.  

\begin{figure}
 \centering
   \includegraphics*[scale=0.5]{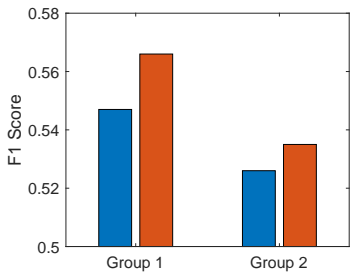}
 \caption{Case Study. Group1: Recommended for Banquets; Group2: Not Recommended for Banquets; Blue: Training with All Instances; Red: Training with Splitted Instances.}
 \label{fig:case-perf}
\end{figure}

We carried out the experiment based on this hypothesis and depicted the results in Figure \ref{fig:case-perf}. Group1 contains the test instances having the feature $b$ and Group2 contains the rest of the test instances. To get the performance showed using the blue bar, we trained a neural network without splitting the training data and calculated F1 scores for Group1 and Group2 separately. As a comparison showed using the red bar, we trained two neural networks with two groups of training data splitted using the aforementioned method. We observe the clear advantage of training and predicting using two different neural networks. This offers compelling evidence for the utility of instance-dependent dropout architecture construction.

\section{Conclusion} \label{sec:conclusion}

In this paper, we proposed a method CODA for instance-dependent dropout architecture construction by applying the human learning theory, constructivism learning to  deep learning. To this end, we proposed a Bayesian
nonparametric method, Uniform Process Mixture Models. This empowers 
our method with the ability to perform assimilation and accommodation, which are two fundamental processes of human
constructivism learning.  The experimental results  show 
that our proposed method has achieved state-of-the-art performance on 
both synthetic data sets and  real-world data sets.

\bibliographystyle{IEEEtran}
\balance
{
\bibliography{dropout}
}

\end{document}